\documentclass[preprint,12pt,number]{elsarticle}
\usepackage{enumitem}
\usepackage{amssymb}
\usepackage{amsmath}

\usepackage[colorlinks=true,linkcolor=black,citecolor=black,urlcolor=blue]{hyperref}

\usepackage{enumitem}
\usepackage{hyperref}
\usepackage{graphicx}

\usepackage{lineno}

\usepackage{graphicx}
\usepackage{subcaption}
\usepackage{array} 
\usepackage{booktabs} 
\usepackage{multirow}   
\usepackage{makecell}
\usepackage{bm}
\usepackage{lineno}

\usepackage{todonotes}

\usepackage[final]{changes}
\usepackage{xcolor}
\usepackage[normalem]{ulem}

\definechangesauthor[name={WL}, color=blue]{WL} 

\newcommand{\wla}{\added[id=WL]}

\journal{Journal of Computational Physics}
\begin{document}

\begin{frontmatter}
\author[aff1,aff2,aff3]{Zhenhua Dang}            
\author[aff1,aff2]{Lei Zhang}                    
\author[aff1,aff2]{Long Wang\corref{cor1}}       
\ead{wanglong@imech.ac.cn}   
\cortext[cor1]{Corresponding author}             
\author[aff1,aff2]{Guowei He\corref{cor2}}       
\ead{hgw@lnm.imech.ac.cn}
\cortext[cor2]{Corresponding author}

\address[aff1]{
  State Key Laboratory of Nonlinear Mechanics, Institute of Mechanics, Chinese Academy of Sciences, Beijing 100190, China
}
\address[aff2]{
  School of Engineering Sciences, University of Chinese Academy of Sciences, 
  Beijing 100049, China
}
\address[aff3]{
  School of Mathematics and Statistics, Northwestern Polytechnical University, 
  Xi'an 710072, China
}

\title{Balance-Guided Sparse Identification of Multiscale Nonlinear PDEs with Small-coefficient Terms}

%% Abstract
\begin{abstract}

Data-driven discovery of governing equations has advanced significantly in recent years; however, existing methods often struggle in multiscale systems where dynamically significant terms may have small coefficients. Therefore, we propose Balance-Guided SINDy (BG-SINDy) inspired by the principle of dominant balance, which reformulates $\ell_0$-constrained sparse regression as a term-level $\ell_{2,0}$-regularized problem and solves it using a progressive pruning strategy. Terms are ranked according to their relative contributions to the governing equation balance rather than their absolute coefficient magnitudes. Based on this criterion, BG-SINDy alternates between least-squares regression and elimination of negligible terms, thereby preserving dynamically significant terms even when their coefficients are small. Numerical experiments on the Korteweg--de Vries equation with a small dispersion coefficient, a modified Burgers equation with vanishing hyperviscosity, a modified Kuramoto--Sivashinsky equation with multiple small-coefficient terms, and a two-dimensional reaction--diffusion system demonstrate the validity of BG-SINDy in discovering small-coefficient terms. The proposed method thus provides an efficient approach for discovering governing equations that contain small-coefficient terms.

\end{abstract}

%%Graphical abstract
\begin{graphicalabstract}
\begin{figure}[htb]
	\centering
	\includegraphics[width=1\linewidth]{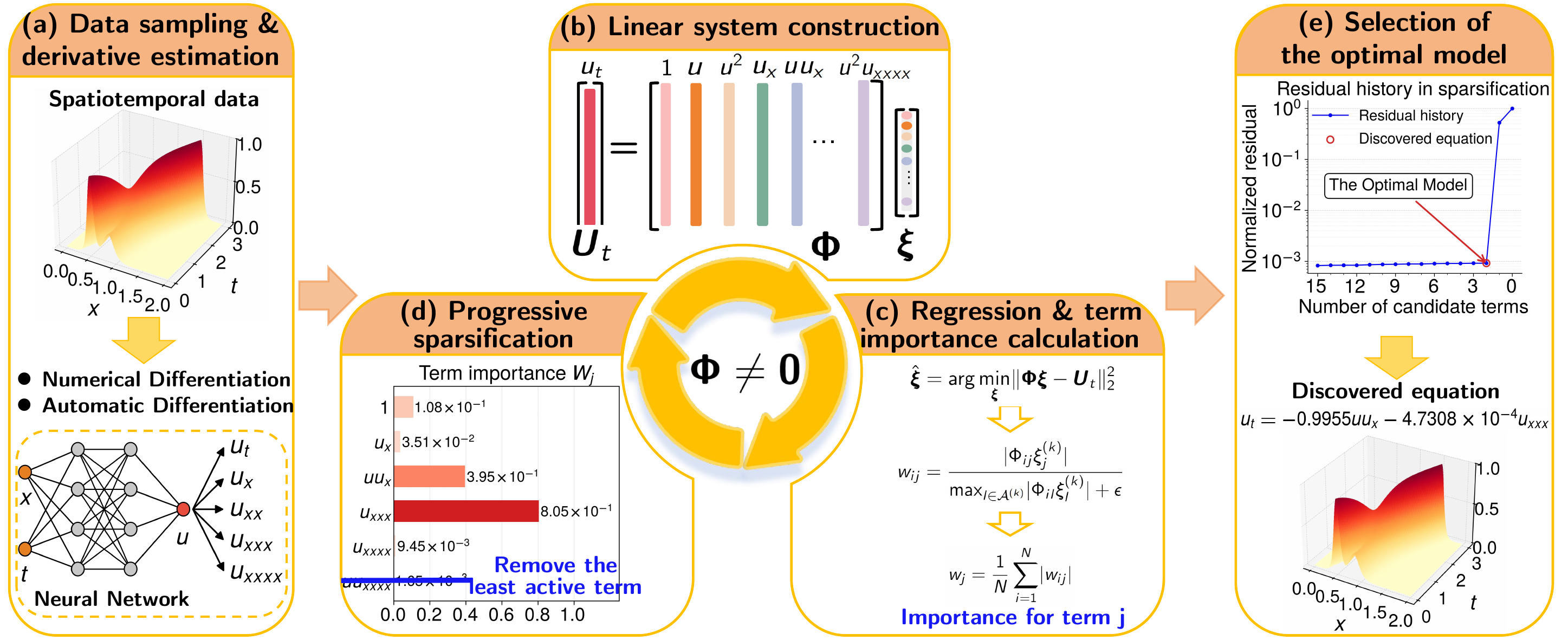}
\end{figure}
\end{graphicalabstract}

%%Research highlights
\begin{highlights}
\item Balance-guided sparsification enables identification of physically important small-coefficient terms.
\item Progressive pruning discards insignificant terms based on physical importance instead of predefined coefficient thresholding.
\item Residual-based model selection identifies the optimal sparse governing equation.
\end{highlights}


\end{frontmatter}

% \linenumbers

\section{Introduction}

Data-driven discovery of governing equations has emerged as a powerful paradigm for uncovering the underlying mechanisms of complex dynamical systems, particularly when first-principles derivations are intractable~\cite{Voss1998continuous,Bar1999fitresidual,Schmidt2009,Josh2007,Brunton2016,Rudy2017,Champion2019}. However, robustly identifying governing equations that contain small-coefficient terms in multiscale systems remains a critical challenge~\cite{Fung2025}. Despite their small magnitudes, such terms cannot be neglected, as they may become locally dominant in specific spatiotemporal regions (e.g., viscous terms in high-Reynolds-number boundary layers~\cite{Prandtl1905,SchlichtingGersten2017} and secular terms in the three-body problem~\cite{MurrayDermott1999}); omitting them disrupts the dominant balance and can lead to fundamentally incorrect models. Nevertheless, most existing equation-discovery methods rely on sparsity-promoting techniques, typically based on practical approximations to $\ell_0$ regularization of coefficients, which select terms according to their coefficient magnitudes~\cite{Brunton2016,Rudy2017,Both2021}. Consequently, small-coefficient terms are often inadvertently discarded during sparsification, even if their \deleted{dynamical contributions} \added{physical contributions} are essential in certain regimes~\cite{Fung2025}. To overcome this sparsity bias, we propose Balance-Guided SINDy (BG-SINDy), inspired by the classical dominant balance principle~\cite{BenderOrszag1999,nayfeh2024perturbation,holmes2012introduction,kevorkian2012multiple}, to discover governing equations for multiscale systems with small-coefficient terms.

Over the past decade, substantial progresses have been achieved in discovering governing equations from observational data through sparse regression and symbolic regression approaches. The sparse identification of nonlinear dynamics (SINDy) framework~\cite{Brunton2016} pioneered sparse regression over libraries of candidate functions and has been extended to partial differential equations via PDE-FIND~\cite{Rudy2017}. To improve robustness against noise and limited data, weak-form formulations~\cite{MessengerBortz2021} and ensemble learning ~\cite{Fasel2022} have been introduced. The widely used Python package PySINDy~\cite{deSilva2020PySINDy} has greatly facilitated the adoption of these methods. Multi-step-accumulation (MSA)-based sparse identification has also been proposed to improve robustness to noisy and incomplete measurements by minimizing the accumulated prediction error over multiple time steps~\cite{Zhang2022multistep}. In addition, the integration of physics-informed neural networks (PINNs)~\cite{raissi2019pinn,Karniadakis2021} has produced hybrid PINN-SINDy~\cite{Chen2021Sun} approaches that enable accurate discovery of governing equations even from scarce or highly noisy measurements. However, PINNs often struggle to accurately identify small perturbation parameters in singularly perturbed systems due to sharp boundary layers and severe loss imbalance; consequently, specialized variants such as the scale-decomposed PINN (SD-PINN)~\cite{zhang2026scale} have been developed to address these specific challenges.

A powerful complement to sparse regression is symbolic regression, which discovers free-form analytical expressions by evolving symbolic structures from elementary operators and variables. A major breakthrough came with the AI Feynman algorithm, which integrates neural-network guidance with brute-force symbolic search~\cite{UdrescuTegmark2020}. This laid the foundation for advances in evolutionary computation, including symbolic genetic algorithms~\cite{Chen2022SGA} and efficient modern implementations such as PySR~\cite{Cranmer2023}, which excel in chaotic and nonlinear dynamical systems. Kolmogorov--Arnold networks (KANs)~\cite{liu2025kan,Zhou2026} have further bridged neural networks and symbolic regression by learning spline-based univariate functions that can be directly translated into explicit symbolic expressions. More recently, large-language-model-assisted approaches have emerged that dynamically update candidate libraries or generate novel expressions~\cite{Du2024LLM}. At the current frontier, generative transformer models such as EqGPT learn co-occurrence patterns of PDE terms directly from mathematical handbooks, enabling autonomous discovery of free-form equations~\cite{Xu2025EqGPT}. Collectively, these methods constitute the primary avenues for interpretable and physics-consistent equation discovery from observational data.

Despite these advances, accurately retaining small-coefficient terms remains a significant challenge in data-driven equation discovery. The SPIDER method~\cite{Gurevich2024} addresses this through residual-driven greedy pruning, which retains candidate terms by monitoring whether their removal causes a sharp increase in the residual. This approach has successfully recovered the Navier--Stokes equations with extremely low viscosity from high-Reynolds-number turbulent channel flow data~\cite{Gurevich2024}, as well as the governing equations of Rayleigh--B\'enard convection~\cite{Wareing2025}, active nematics~\cite{Golden2023}, and turbulent magnetohydrodynamic flows~\cite{Golden2025}. The SPRINT algorithm~\cite{Golden2024} accelerates the pruning process using bisection with analytic bounds on rank-1 SVD updates while preserving sensitivity to minute coefficients even at large library sizes. Similarly, the bi-level optimization framework BILLIE~\cite{Li2025BILLIE} decouples term selection from coefficient estimation via reinforcement learning and independent validation splits, thereby preserving subtle yet dynamically essential contributions even under noise or data scarcity. Despite their notable success, these methods typically incur substantial computational overhead due to exhaustive greedy searches, large-library SVD computations, or bi-level reinforcement learning.

To provide a computationally efficient and physically principled alternative, we draw inspiration from the dominant balance principle~\cite{BenderOrszag1999,nayfeh2024perturbation,holmes2012introduction,kevorkian2012multiple}, a cornerstone of asymptotic analysis in complex multiscale systems. By systematically comparing the relative magnitudes of terms in the governing equations, this principle identifies the dominant \deleted{dynamical contributions} \added{physical contributions} under specific limiting regimes. A classic example is the derivation of the Prandtl boundary-layer equations from the full Navier--Stokes equations in the high-Reynolds-number limit~\cite{Prandtl1905}. Complementary techniques, such as matched asymptotic expansions~\cite{Hinch1991}, multiple-scale analysis~\cite{KevorkianCole1996}, and the WKB approximation~\cite{BenderOrszag1999}, enable the systematic reduction of complex models into their essential dynamical structures. More recently, this paradigm has been extended to data-driven reduced-order models~\cite{Callaham2021} when the complete governing equations are known \emph{a priori}. Using unsupervised learning techniques such as Gaussian mixture models and sparse principal component analysis, local dominant balances can be identified as distinct clusters in an ``equation space'', thereby automatically determining which terms may be safely neglected in different spatiotemporal regions without compromising model fidelity.

In this paper, we propose Balance-Guided SINDy (BG-SINDy), a novel sparse regression framework for discovering governing equations of multiscale dynamical systems. Inspired by the principle of dominant balance, BG-SINDy reformulates the conventional coefficient-level $  \ell_0  $ regularization as a term-level $  \ell_{2,0}  $ problem and solves it via a progressive pruning strategy. By prioritizing the relative physical contributions of candidate terms to the overall equation balance rather than their magnitudes, the method iteratively alternates between constrained least-squares estimation and importance-based pruning. This enables accurate discovery of parsimonious models that retain essential small-coefficient terms in multiscale systems. The main contributions of this work are summarized as follows:
\begin{enumerate}[label=(\roman*)]

\item A balance-guided sparse regression framework that reformulates coefficient-level $\ell_0$ regularization as term-level $\ell_{2,0}$ regularization.
\item A progressive pruning strategy that sequentially eliminates candidate terms according to their relative physical contributions, thereby avoiding heuristic coefficient thresholds.
\item A residual-based criterion that automatically selects the optimal sparse representation of the governing equation.
\end{enumerate}

The remainder of this paper is organized as follows. Section~\ref{sec:2} presents the mathematical formulation and algorithmic implementation of the BG-SINDy framework. Section~\ref{sec:3} validates the effectiveness of the proposed method on challenging cases, including high-order derivative terms with small coefficients, multiple small-coefficient terms, and scalability to high-dimensional systems. Concluding remarks and future directions are provided in Section~\ref{sec:4}.

\section{Method~\label{sec:2}}
In this section, we first review the SINDy method and then present the proposed BG-SINDy method. For simplicity, the following introduction of the two methods is illustrated using one-dimensional examples.

\subsection{SINDy}\label{subsec:SINDy}
In the SINDy framework, the evolution of the state variable \(u(x,t)\) is assumed to satisfy
\begin{equation}
\frac{\partial u}{\partial t} = \boldsymbol{\varphi}(u)\cdot \boldsymbol{\xi},
\end{equation}
where \(\boldsymbol{\varphi}(u)\) is a row vector of candidate nonlinear functions constructed from polynomials of \(u\) and its spatial derivatives, and \(\boldsymbol{\xi}\) is the coefficient vector. The candidate library consists of polynomial terms up to degree \(P\) and their products with spatial derivatives up to order \(Q\):
\[
\left\{ u^p \;\middle| p=0,\dots,P \right\}
\cup
\left\{ u^p \frac{\partial^q u}{\partial x^q} \;\middle| p=0,\dots,P;\ q=1,\dots,Q \right\}.
\]
Accordingly, the total number of candidate terms is \(M = (P+1)(Q+1)\).

Evaluating these terms at sampled spatiotemporal points \((x_i,t_k)\) yields the overdetermined regression system
\begin{equation}
\mathbf{U}_t = \boldsymbol{\Phi}\boldsymbol{\xi},
\end{equation}
where \(\mathbf{U}_t \in \mathbb{R}^{N \times 1}\) collects the time derivatives at \(N\) sampling points, and \(\boldsymbol{\Phi} \in \mathbb{R}^{N \times M}\) is the library matrix. Each row of \(\boldsymbol{\Phi}\) contains the evaluated candidate terms at a given point, with \(\Phi_{ij}\) denoting the value of the \(j\)-th term at the \(i\)-th sample, and \(\boldsymbol{\xi} \in \mathbb{R}^{M \times 1}\). The required derivatives can be approximated either by classical numerical methods (finite differences, spectral methods, etc.)~\cite{Rudy2017, Fukami2021Sparse, Kaheman2020SINDyPI} or by automatic differentiation through neural-network surrogates~\cite{Baydin2018ADSurvey, Xu2021DLPDE, Kaheman2022Automatic}.

The sparse vector \( \boldsymbol{\xi} \) is then recovered by solving the \( \ell_0 \)-regularized least-squares problem~\cite{Brunton2016}
\begin{equation}\label{eq:SINDy}
\hat{\boldsymbol{\xi}} = \arg\min_{\boldsymbol{\xi}} \ \|\mathbf{U}_t - \boldsymbol{\Phi}\boldsymbol{\xi}\|_2^2 + \lambda \|\boldsymbol{\xi}\|_0,
\end{equation}
where \( \lambda > 0 \) controls the level of sparsity, and \(\|\boldsymbol{\xi}\|_0\) counts the non-zero elements in the coefficient vector \(\boldsymbol{\xi}\). In practice, this NP-hard problem~\cite{Natarajan1995} is approximately solved using sparsity-promoting algorithms within the SINDy framework. The original SINDy algorithm~\cite{Brunton2016} employs the sequential thresholded least-squares (STLSQ) method to discover ordinary differential equations (ODEs), which iteratively alternates between ordinary least-squares regression and hard thresholding of small coefficients. For the discovery of partial differential equations (PDEs), the PDE-FIND algorithm~\cite{Rudy2017}, a popular extension of the SINDy framework, adopts the TrainSTRidge method (train sequential thresholded ridge regression), which combines ridge regression with sequential thresholding. The threshold is selected via a train-validation split to improve robustness to noise.

\subsection{BG-SINDy}
The BG-SINDy method promotes sparsity at the level of candidate terms according to their contributions to the equation balance evaluated at sampling points, rather than at the individual coefficient level, in the spirit of the principle of dominant balance. The optimal sparse governing equation is obtained by solving
\begin{equation}\label{eq:BG-SINDy}
\hat{\boldsymbol{\xi}} = \arg\min_{\boldsymbol{\xi}}
\|\mathbf{U}_t - \boldsymbol{\Phi}\boldsymbol{\xi}\|_2^2
+ \lambda \bigl\|\operatorname{diag}(\boldsymbol{\xi})\boldsymbol{\Phi}^{T}\bigr\|_{2,0},
\end{equation}
where \(\mathbf{U}_t \in \mathbb{R}^{N\times 1}\), \(\boldsymbol{\Phi} \in \mathbb{R}^{N\times M}\), and \(\boldsymbol{\xi} \in \mathbb{R}^{M\times 1}\) follow the definitions in Sec.~\ref{subsec:SINDy}. Here, \(\operatorname{diag}(\boldsymbol{\xi})\) is a diagonal matrix formed from \(\boldsymbol{\xi}\), and \(\|\cdot\|_{2,0}\) counts the number of nonzero rows of a matrix~\cite{cai2013exact,pang2018efficient}. 

Since each nonzero row of $  \operatorname{diag}(\boldsymbol{\xi})\boldsymbol{\Phi}^{T}  $ corresponds uniquely to one active candidate term, Eq.~\eqref{eq:BG-SINDy} reformulates the coefficient-level $\ell_0$ regularization as an equivalent term-level $ \ell_{2,0}  $ regularization. In particular, \(\bigl\|\operatorname{diag}(\boldsymbol{\xi})\boldsymbol{\Phi}^{T}\bigr\|_{2,0}\) counts the number of nonzero columns of \(\boldsymbol{\Phi}\), i.e., the number of active candidate terms. This formulation enforces sparsity directly at the term-level, enabling entire candidate functions to be selected or discarded based on their collective spatiotemporal contribution. This shifts the sparsity mechanism from coefficient magnitude to \added{term-level contribution}, fundamentally altering the model selection criterion. In contrast to standard SINDy, which relies on coefficient thresholding, BG-SINDy selects terms according to their \added{dynamical importance}, consistent with the dominant balance principle~\cite{BenderOrszag1999,nayfeh2024perturbation,holmes2012introduction,kevorkian2012multiple}.

\begin{figure}[t!]
\centering
\includegraphics[width=1.0\linewidth]{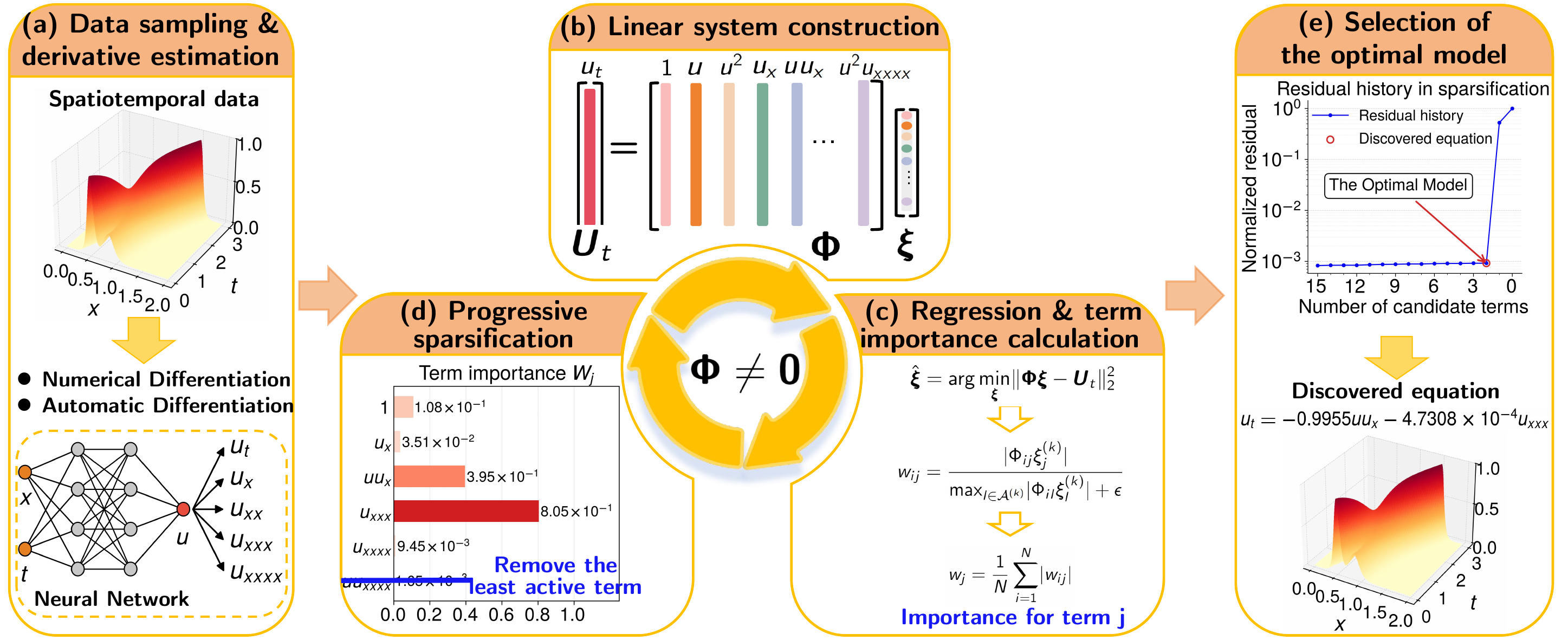}
\caption{Schematic of the BG-SINDy method. (a) Spatio-temporal data sampling and derivative estimation via numerical differentiation or automatic differentiation.
(b) Construction of the linear system $\mathbf{U}_t=\boldsymbol{\Phi} \boldsymbol{\xi}$.
(c) Regression and term importance calculation.
(d) Progressive pruning by iteratively discarding the least significant terms.
(e) Identification of the optimal sparse model based on the evolution of residuals.
}
\label{fig:framework_PTBSINDy}
\end{figure}

To solve the $l_{2,0}$-norm constrained regression problem in Eq.~\eqref{eq:BG-SINDy} effectively, we propose a progressive pruning workflow that realizes Occam's razor~\cite{Champion2019,Lemos2022Rediscovering} through term-level sparsification. As illustrated in Fig.~\ref{fig:framework_PTBSINDy}, the procedure proceeds through the following iterative steps:

\begin{enumerate}[label=(\roman*)]
\item \textit{Spatio-temporal data sampling and derivative estimation}
(Fig.~\ref{fig:framework_PTBSINDy}(a)): We first collect spatio-temporal data and then compute the required derivatives accurately using either numerical differentiation or automatic differentiation~\cite{Baydin2018ADSurvey,Kaheman2022Automatic}, for example via PyTorch-based neural networks~\cite{paszke2019pytorch}. The resulting dataset consists of $N$ spatio-temporal sample points, forming the target time-derivative vector $\mathbf{U}_t \in \mathbb{R}^N$.
\par\noindent
\item \textit{Linear system construction}
(Fig.~\ref{fig:framework_PTBSINDy}(b))): Based on the state variables, an initial library of candidate terms is constructed. Progressive pruning assumes that no linear combination of candidate terms constitutes a valid equation on its own. This linear independence prevents term coupling and enables the individual elimination of redundant terms. If the initial library violates this, a maximal linearly independent subset is extracted \emph{a priori} using singular value decomposition or QR factorization with column pivoting~\cite{Golub2013} to construct the candidate library matrix $\boldsymbol{\Phi} \in \mathbb{R}^{N\times M}$. We initialize the active set as $\mathcal{A}^{(0)} = \{1,2,\dots,M\}$ and set the iteration counter $k = 0$. The initial least-squares estimate is computed as:
\begin{equation}
\boldsymbol{\xi}^{(0)} 
= 
\arg\min_{\boldsymbol{\xi}} 
\left\|
\boldsymbol{\Phi}_{:,\mathcal{A}^{(0)}} \boldsymbol{\xi} - \mathbf{U}_t
\right\|_2^2,
\end{equation}
with the initial \deleted{residual}
\added{residual of the fitting model} defined as:
\begin{equation}
\mathrm{Res}_0
=
\frac{1}{N}
\left\|
\boldsymbol{\Phi}_{:,\mathcal{A}^{(0)}} \boldsymbol{\xi}^{(0)} - \mathbf{U}_t
\right\|_2^2.
\end{equation}

\item \textit{Regression and term importance calculation}
(Fig.~\ref{fig:framework_PTBSINDy}(c)): At iteration $k$, given the regression coefficients $\boldsymbol{\xi}^{(k)}$ on the current active set $\mathcal{A}^{(k)}$, we quantify the contribution of each term to the equation balance in the same spirit as the principle of dominant balance. For each sample point $i=1,\dots,N$ and each active term $j\in\mathcal{A}^{(k)}$, the local relative importance is defined as
\begin{equation}\label{eq:local-importance}
w_{ij}
=
\frac{\left|\Phi_{ij}\,\xi^{(k)}_j\right|}
{\max_{l \in \mathcal{A}^{(k)}} |\Phi_{il}\,\xi^{(k)}_l|+\epsilon},
\end{equation}
where $\epsilon>0$ is a small constant for numerical stability. It follows that $0 \leq w_{ij} \leq 1$.

\item \textit{Progressive pruning}
(Fig.~\ref{fig:framework_PTBSINDy}(d)): The local relative importance is averaged over all sample points to obtain the global importance:
\begin{equation}\label{eq:global-importance}
W_j
=
\frac{1}{N}
\sum_{i=1}^N w_{ij}.
\end{equation}
Consequently, $0 \leq W_j \leq 1$. The least important term is identified as
\[
j^* = \arg\min_{j \in \mathcal{A}^{(k)}} W_j
\]
and removed from the active set:
\[
\mathcal{A}^{(k+1)} \leftarrow \mathcal{A}^{(k)} \setminus \{j^*\}.
\]
The regression coefficients are then updated on the reduced active set by solving the unconstrained least-squares problem
\begin{equation}
\boldsymbol{\xi}^{(k+1)}
=
\arg\min_{\boldsymbol{\xi}}
\left\|
\boldsymbol{\Phi}_{:,\mathcal{A}^{(k+1)}} \boldsymbol{\xi} - \mathbf{U}_t
\right\|_2^2.
\end{equation}
The residual of the fitted model is given by
\begin{equation}
\mathrm{Res}_{k+1}
=
\frac{1}{N}
\left\|
\boldsymbol{\Phi}_{:,\mathcal{A}^{(k+1)}} \boldsymbol{\xi}^{(k+1)} - \mathbf{U}_t
\right\|_2^2.
\end{equation}

\item \textit{Selection of the Optimal Model}
(Fig.~\ref{fig:framework_PTBSINDy}(e)): As terms are progressively removed, the residual typically remains approximately constant until a dynamically essential term is eliminated, at which point it exhibits a substantial increase. After computing $\mathrm{Res}_{k+1}$, the termination condition is evaluated: 
\begin{equation}\label{eq:termination}
\frac{\mathrm{Res}_{k+1}}{\mathrm{Res}_{k}} > \tau,
\end{equation}
where $\tau = 3$ in this work. When this condition is satisfied, the current model is deemed to have reached an overly sparse representation, and the optimal model is taken from the previous iteration corresponding to the step immediately before the increase in the residual:
\[
\boldsymbol{\xi}^* = \boldsymbol{\xi}^{(k)}, 
\quad
\mathcal{A}^* = \mathcal{A}^{(k)}.
\]
Otherwise, the iteration counter is updated as $k \leftarrow k+1$, and the algorithm returns to step (iii).

\end{enumerate}

By transforming the combinatorial \(\ell_{2,0}\)-regularized regression problem into a deterministic iterative procedure guided by physically interpretable criteria of dynamical contribution, the proposed BG-SINDy method offers several key advantages. It enables robust identification of governing equations from observational data while successfully preserving dynamically essential terms with small coefficients. Furthermore, unlike traditional sparsity-promoting techniques that require delicate manual tuning of absolute coefficient thresholds, BG-SINDy systematically identifies the optimal model structure through a residual-change criterion. This fundamentally enhances the physical interpretability of the identified governing equations and facilitates more faithful recovery of underlying dynamics. 

\section{Results~\label{sec:3}}
\wla{We demonstrate the performance of the BG-SINDy method on four benchmark problems for discovering nonlinear governing equations containing small-coefficient terms. The method is first evaluated on the Korteweg--de Vries (KdV) equation with a small dispersion coefficient to assess its robustness to measurement noise and varying numbers of sampling points. Next, a modified Burgers equation augmented with vanishing hyperviscosity is then considered to assess its ability to identify high-order derivative terms with small coefficients. Then, the method's capability to handle equations with multiple small-coefficient terms is examined using a modified Kuramoto--Sivashinsky (KS) equation. Finally, the method's scalability is investigated using a two-dimensional reaction--diffusion equation.}

\subsection{KdV equation with a small dispersion coefficient}

In this example, the KdV equation is used to evaluate the performance of the proposed BG-SINDy method, focusing on its ability to identify terms associated with small coefficients. The KdV equation~\cite{kdV1895} describes the propagation of nonlinear dispersive waves in one spatial dimension and has been widely used in equation discovery studies~\cite{Rudy2017,raissi2019pinn,XuZhang2021RobustPDE,StephanyEarls2024_WeakPDELEARN}. Here we consider the KdV equation with a small dispersion coefficient~\cite{DagDereli2008}
\begin{equation}\label{eq:KdV}
\frac{\partial u}{\partial t} + u\frac{\partial u}{\partial x} + \varepsilon \frac{\partial^3 u}{\partial x^3} = 0, \qquad (x,t)\in \Omega\times[0,T],
\end{equation}
where \(\varepsilon = 4.84 \times 10^{-4}\), \(\Omega = [0,2]\), and \(T=3\). The initial condition is
\[
u(x,0) = 0.9\,\operatorname{sech}^2\!\bigl[12.45(x-0.5)\bigr] + 0.3\,\operatorname{sech}^2\!\bigl[7.1875(x-0.85)\bigr], \quad x\in \Omega,
\]
and homogeneous Dirichlet boundary conditions are imposed.

The reference dataset is generated by numerically solving the KdV equation~\eqref{eq:KdV} using a fourth-order Runge--Kutta scheme in time combined with a finite-difference discretization in space~\cite{KassamTrefethen2005_FourthOrderTimeStepping}. The spatial domain is discretized into \(N_{\Omega}=260\) uniform grid points, and the time step is \(\Delta t=0.001\).

To discover the governing equation from sparse and noisy data, the required derivatives are estimated via automatic differentiation~\cite{Baydin2018ADSurvey,Xu2021DLPDE,Kaheman2022Automatic} on a neural-network surrogate (a fully connected network with five hidden layers of 50 neurons each and sin activation) trained on only 500 points (0.064\% of the full dataset) obtained via Latin hypercube sampling~\cite{mckay1979Latinhyper}. The candidate library consists of polynomial terms in \(u\) up to degree 2, together with their products with spatial derivatives of \(u\) up to order 4, yielding a total of 15 candidate terms:
\[
\left\{ u^p \;\middle|\; p=0,1,2 \right\}
\;\cup\;
\left\{ u^p \frac{\partial^q u}{\partial x^q} \;\middle|\; p=0,1,2;\ q=1,\dots,4 \right\}.
\]

\begin{figure}[ht!]
\centering
\includegraphics[width=1.0\textwidth]{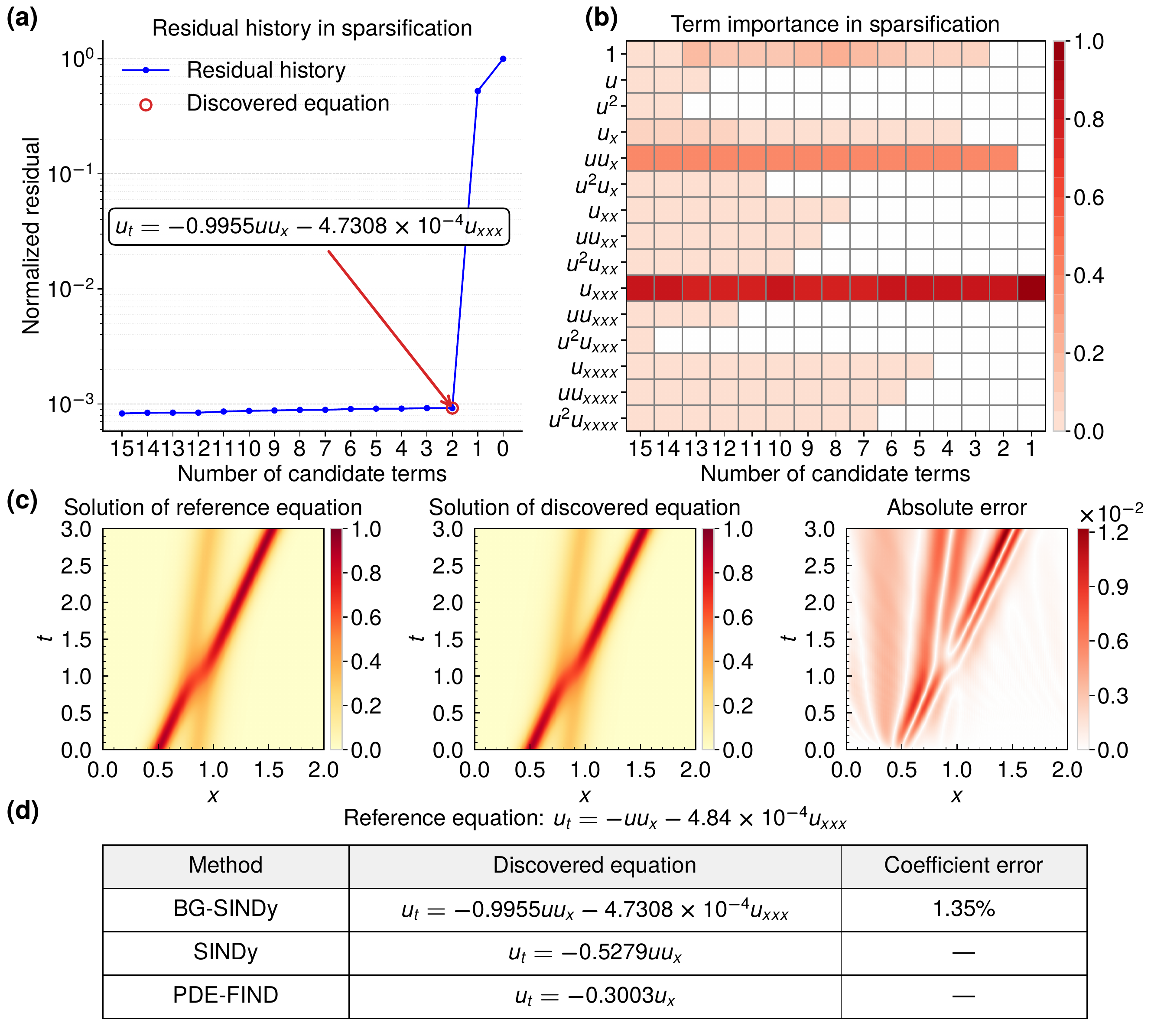}
\caption{Results for the KdV equation. (a) History of residual in the sparsification process. The red circle denotes the discovered equation.
(b) Heatmap of term importance in the sparsification process.
(c) Comparison of the solutions obtained from the reference equation and the discovered equation. The relative $L_2$ error is $1.19\%$.
(d) Discovered equation by BG-SINDy and comparison with SINDy and PDE-FIND. BG-SINDy accurately recovers the reference KdV equation, while SINDy and PDE-FIND fail.}
\label{fig:KdV}
\end{figure}

The performance of BG-SINDy on the KdV equation is summarized in Fig.~\ref{fig:KdV}. Following the workflow illustrated in Fig.~\ref{fig:framework_PTBSINDy}, the candidate library is progressively pruned by iteratively eliminating the least significant term according to the \added{importance score} defined in Eqs.~\eqref{eq:local-importance} and~\eqref{eq:global-importance}. As shown in Fig.~\ref{fig:KdV}(a), this procedure yields a sequence of increasingly sparse models accompanied by a monotonic increase in the residual. The selected model, indicated by the red circle, corresponds to the elbow point (pronounced change) in the residual. The evolution of \added{term importance} during the sparsification process is visualized in the heatmap in Fig.~\ref{fig:KdV}(b).
The validity of the discovered equation is further verified by numerically integrating the model and comparing the resulting solution with the reference solution shown in Fig.~\ref{fig:KdV}(c), yielding a relative $  L_2  $ error of $1.19\%$. A comparison with SINDy and PDE-FIND under the same candidate library and derivative estimation is shown in Fig.~\ref{fig:KdV}(d). When the equation form is correctly identified, accuracy is quantified by the coefficient error~\cite{Xu2025EqGPT}: 
\begin{equation}\label{eq:coeff}
  \text{coefficient error} = \frac{1}{N_{\text{term}}} \sum_{i=1}^{N_{\text{term}}} \frac{|\xi_i - \xi_i^{\text{true}}|}{|\xi_i^{\text{true}}|},   
\end{equation}
where \(N_{\rm term}\) is the number of terms that appear in both the discovered and reference equations, and \(\xi_i\), \(\xi_i^{\rm true}\) are the corresponding coefficients in the discovered and reference models, respectively. BG-SINDy successfully recovers the KdV equation, with a coefficient error of only $1.35\%$, whereas both SINDy and PDE-FIND fail to identify the equation structure. 

\begin{figure}[ht!]
\centering
\includegraphics[width=0.6\textwidth]{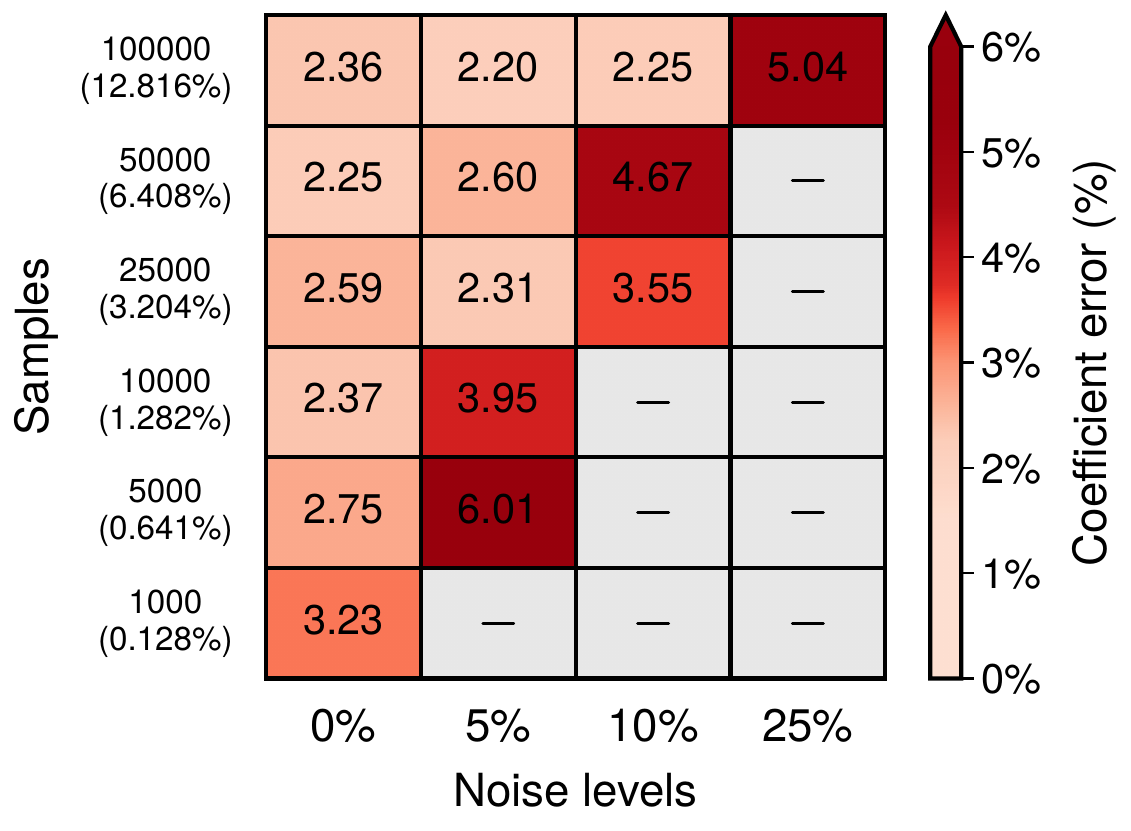}
\caption{Heatmap of the coefficient error in the BG-SINDy discovery of the KdV equation across different noise levels and sampling data sizes, computed according to Eq.~\eqref{eq:coeff}. The colormap ranges from light red (low error) to dark red (high error). Gray cells indicate failed discovery of the correct equation structure.}
\label{kdv_datarobust_figure}
\end{figure}

The robustness of BG-SINDy to noise and sampling data size is assessed by adding Gaussian white noise to the clean data following~\cite{Rudy2017,Xu2025EqGPT}:
$$\tilde{u} = u + \gamma \cdot \mathrm{std}(u) \cdot \mathcal{N}(0,1),$$
where $  \gamma  $ denotes the noise amplitude. The coefficient error, computed according to Eq.~\eqref{eq:coeff} for noise levels ranging from $0\%$ to $25\%$ and sampling sample sizes from $1{,}000$ to $100{,}000$, is presented in the heatmap of Fig.~\ref{kdv_datarobust_figure}. Even at $25\%$ noise, BG-SINDy maintains low coefficient errors when sufficient data are available.

\subsection{Modified Burgers equation with vanishing hyperviscosity}

This example aims to highlight BG-SINDy's ability to discover governing equations that contain small-coefficient terms involving high-order derivatives. We consider the modified Burgers equation augmented with a vanishing hyperviscosity term~\cite{Tadmor2004,Frisch2008Hyperviscosity} (i.e., a fourth-order derivative with a small coefficient):
\begin{equation}
\begin{split}
\frac{\partial u}{\partial t} + u \frac{\partial u}{\partial x}
- 0.5 \frac{\partial^2 u}{\partial x^2}
+ \varepsilon \frac{\partial^4 u}{\partial x^4} &= 0,
\qquad (x,t)\in \Omega\times[0,T],
\end{split}
\label{eq:KS}
\end{equation}
subject to the initial condition
$$   u(x,0) = \cos\left(\frac{x}{16}\right), \quad x \in \Omega,   $$
and periodic boundary conditions, where $  \varepsilon = 1 \times 10^{-3}  $, $  \Omega = [0,32\pi]  $, and $  T = 100  $. 
The dataset is generated by numerically solving Eq.~\eqref{eq:KS} using a fourth-order Runge--Kutta integrator with time step $  \Delta t = 0.1  $ and Fourier spectral discretization with 4048 modes~\cite{MessengerBortz2021,KassamTrefethen2005_FourthOrderTimeStepping}. Temporal derivatives are computed via finite differences, while spatial derivatives are evaluated spectrally. The candidate library comprises the polynomial terms $u^p$ ($p=0,1,2$) and all products of these terms with spatial derivatives $\frac{\partial^q u}{\partial x^q}$($q=1,\dots,4$), yielding a total of 15 terms:
\[
\left\{ u^p \;\middle|\; p=0,1,2 \right\}
\;\cup\;
\left\{ u^p \frac{\partial^q u}{\partial x^q} \;\middle|\; p=0,1,2;\quad q=1,\dots,4 \right\}.
\]

\begin{figure}[ht!]
\centering
\includegraphics[width=1.0\textwidth]{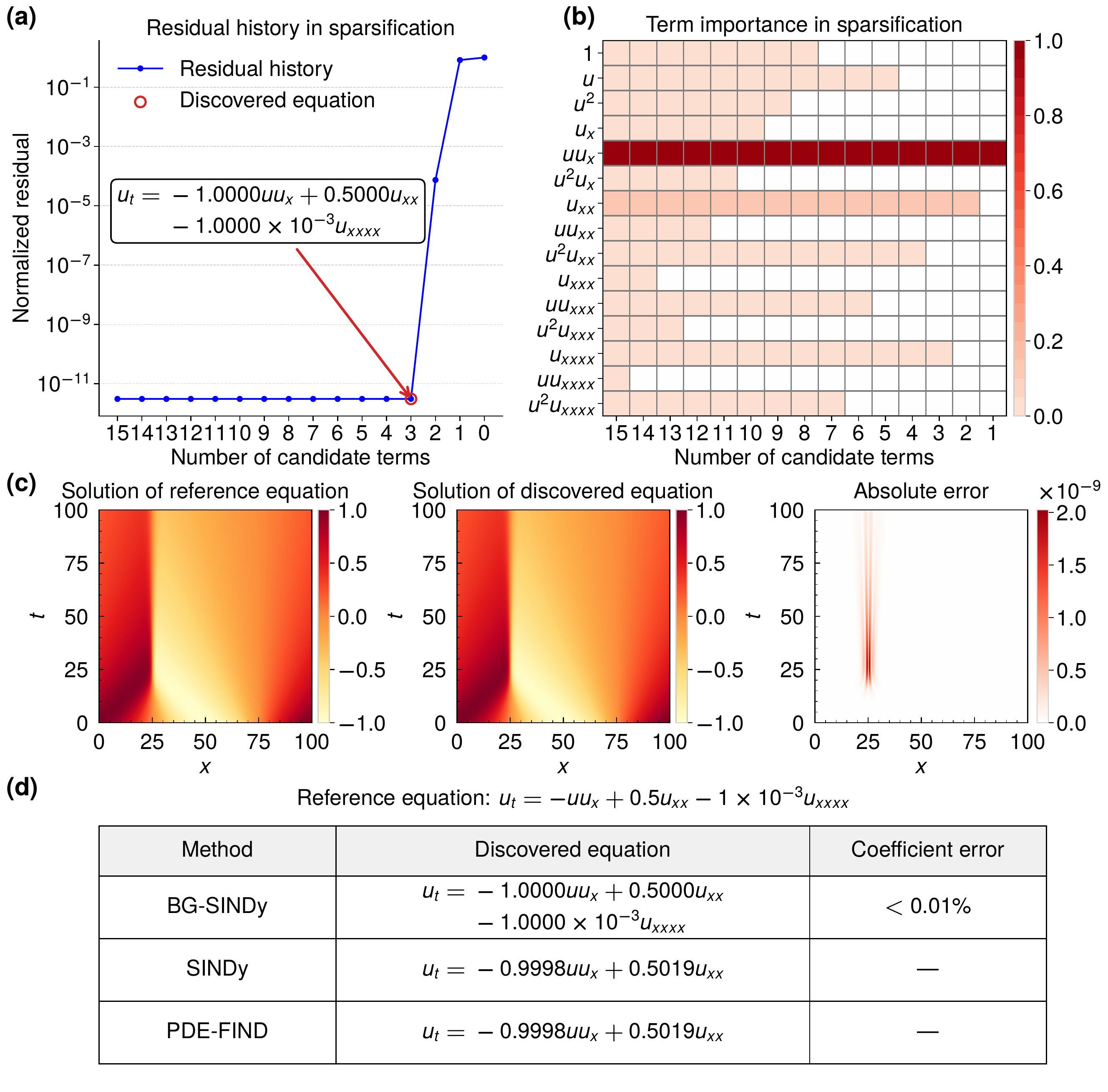}
\caption{Results for the modified Burgers equation. 
(a) History of residual in the sparsification process. The red circle denotes the discovered equation.
(b) Heatmap of term importance in the sparsification process.
(c) Comparison of the solutions obtained from the reference equation and the discovered equation. The relative $L_2$ error is less than $0.01\%$.
(d) Discovered equation by BG-SINDy and comparison with SINDy and PDE-FIND. BG-SINDy accurately recovers the reference equation, while SINDy and PDE-FIND fail to identify the small-coefficient term.}
\label{fig:Burger-hyper-vis}
\end{figure}

The performance of BG-SINDy on the modified Burgers equation with vanishing hyperviscosity is summarized in Fig.~\ref{fig:KdV}. As shown in Fig.~\ref{fig:Burger-hyper-vis}(a), the residual increases monotonically as the candidate library is pruned. The selected model, marked by the red circle, corresponds to the elbow point, indicating a pronounced change in the residual. The evolution of term importance during the sparsification process is visualized in the heatmap in Fig.~\ref{fig:Burger-hyper-vis}(b).
The validity of the discovered equation is further verified by numerically integrating the model and comparing the resulting solution with the reference solution shown in Fig.~\ref{fig:Burger-hyper-vis}(c), yielding a relative $  L_2  $ error below $0.01\%$. A comparison with SINDy and PDE-FIND under the same candidate library and derivative estimation is shown in Fig.~\ref{fig:Burger-hyper-vis}(d). 
BG-SINDy accurately recovers the modified Burgers equation with a coefficient error below $0.01\%$, whereas both SINDy and PDE-FIND identify only the nonlinear and viscous terms and miss the small-coefficient hyperviscosity term. This example demonstrates the robustness of BG-SINDy to small-coefficient high-order derivative terms.  

\subsection{Modified KS equation with multiple small-coefficient nonlinear terms}
This example aims to highlight the ability of BG-SINDy to discover governing equations containing multiple small-coefficient terms from data. We consider the modified KS equation augmented with multiple small-coefficient nonlinear terms~\cite{Golden2024}
\begin{equation}
\begin{split}
\frac{\partial u}{\partial t} + u \, \frac{\partial u}{\partial x} + \frac{\partial^2 u}{\partial x^2} + \frac{\partial^4 u}{\partial x^4} + \varepsilon \sum_{k=3}^{6} \frac{\partial}{\partial x} \left( u^k \right) &= 0,
\qquad (x,t)\in \Omega\times[0,T],
\end{split}
\label{eq:modKS}
\end{equation}
with the initial condition
\[
u(x,0) = \cos(3x) - \tfrac{1}{2}\sin(x), 
\quad x \in \Omega .
\]
and the periodic boundary condition, where \(\varepsilon = 1\times 10^{-6}\) , \(\Omega = [0,22]\) and \(T = 200\).

The candidate library comprises polynomial terms in $u$ up to degree 10, together with their products with spatial derivatives of $u$ up to 10th order, resulting in 121 candidate terms:
\[
\left\{ u^p \;\middle|\; p=0,1,\dots,10 \right\}
\;\cup\;
\left\{ u^p \frac{\partial^q u}{\partial x^q} \;\middle|\; p=0,1,\dots,10;\quad q=1,\dots, 10 \right\}.
\] 
These terms are constructed from the solution $u$ obtained by numerically solving the modified KS equation~\eqref{eq:modKS} using a sixth-order Runge--Kutta scheme with time step $\Delta t = 0.004$ and Fourier spectral discretization with 128 modes~\cite{Golden2024, Butcher1964Implicit}. Temporal derivatives are computed using the finite difference method, and spatial derivatives are computed using the spectral method.

\begin{figure}[htbp!]
\centering
\includegraphics[width=1.0\textwidth]{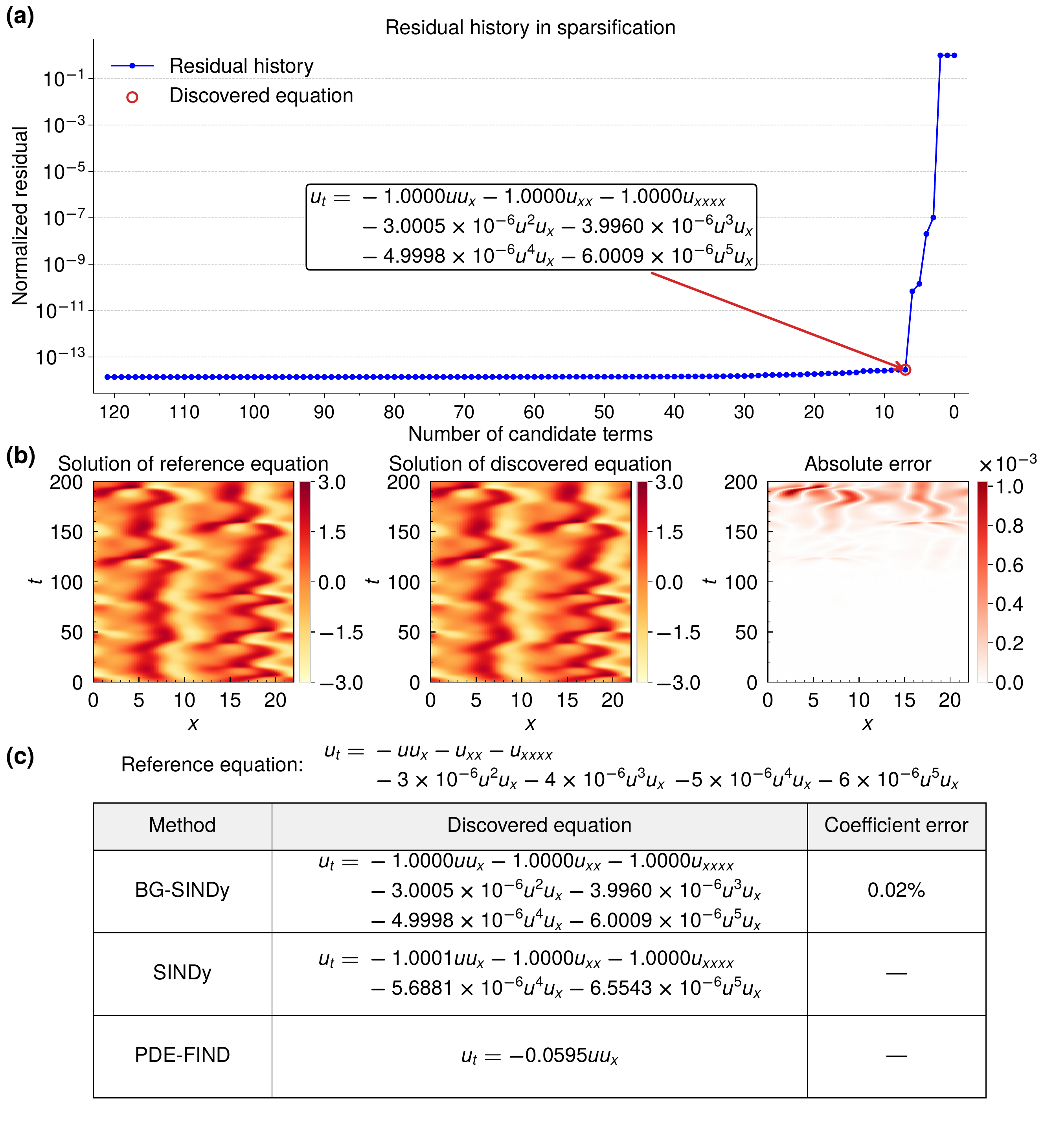}
\caption{Results for the modified KS equation. 
(a) History of residual in the sparsification process. The red circle denotes the discovered equation.
(b) Comparison of the solutions obtained from the reference equation and the discovered equation. The relative $L_2$ error below  $0.01\%$.
(c) Discovered equation by BG-SINDy and comparison with SINDy and PDE-FIND. BG-SINDy accurately recovers the reference modified KS equation, whereas SINDy only recovers the two terms with small coefficients, and PDE-FIND fails to identify any of the small-coefficient terms.}
\label{fig:ModBurgers}
\end{figure}

The performance of BG-SINDy on the modified KS equation with multiple small-coefficient nonlinear terms is summarized in Fig.~\ref{fig:ModBurgers}. As shown in Fig.~\ref{fig:ModBurgers}(a), the residual increases monotonically as the candidate library is pruned. The selected model, marked by the red circle, corresponds to the elbow point, indicating a pronounced change in the residual. The validity of the discovered equation is further verified by numerically integrating the model and comparing the resulting solution with the reference solution shown in Fig.~\ref{fig:ModBurgers}(b), yielding a relative $  L_2  $ error below $0.01\%$. A comparison with SINDy and PDE-FIND under the same candidate library and derivative estimation is shown in Fig.~\ref{fig:ModBurgers}(c). BG-SINDy accurately recovers the modified KS equation with a coefficient error of $0.02\%$. In contrast, SINDy only recovers the two terms with small coefficients, and PDE-FIND fails to identify any of the small-coefficient terms. This example demonstrates the robustness of BG-SINDy to multiple small-coefficient terms.

\subsection{Two-dimensional reaction--diffusion system}

This example aims to highlight the ability of BG-SINDy to discover coupled governing equations in two spatial dimensions. The two-dimensional reaction--diffusion system~\cite{Turing1952,Murray2003} is a widely studied model that exhibits complex pattern formation through the interplay of nonlinear reaction kinetics and diffusion. Here we consider the reaction--diffusion equation with small diffusion coefficients
\begin{equation}
\begin{split}
\frac{\partial u}{\partial t} &= u+0.5 v^3 - uv^2 + 0.5 u^2 v - u^3  
+ \varepsilon \left( \frac{\partial^2 u}{\partial x^2} + \frac{\partial^2 u}{\partial y^2} \right), \\
\frac{\partial v}{\partial t} &= v- v^3 - 0.5 u v^2 - u^2 v - 0.5 u^3  
+ \varepsilon \left( \frac{\partial^2 v}{\partial x^2} + \frac{\partial^2 v}{\partial y^2} \right), \\
&  \hfill \hspace{7cm} (x,y,t) \in \Omega \times [0,T],
\end{split}
\label{eq:2dRD}
\end{equation}
where $\Omega = [-1.5,1.5] \times [-1.5,1.5]$, $\varepsilon = 1 \times 10^{-3}$, and $T = 5$. The initial conditions are given by
\begin{equation} 
\begin{aligned}
u(x,y,0)&=\tanh\!\big(\sqrt{x^2+y^2}\big)\cos\!\left(2\theta(x+\mathrm{i}y)-\sqrt{x^2+y^2}\right),
\\
v(x,y,0)&=\tanh\!\big(\sqrt{x^2+y^2}\big)\sin\!\left(2\theta(x+\mathrm{i}y)-\sqrt{x^2+y^2}\right),
\end{aligned}
\end{equation}
where $\theta(z)$ is the principal angle of $z\in\mathbb{C}$, and periodic boundary conditions are imposed~\cite{deSilva2020PySINDy}.

The reference dataset is generated by numerically solving the 2D reaction--diffusion system~\eqref{eq:2dRD} using the adaptive RK45 solver~\cite{DormandPrince1980,Virtanen2020SciPy} in time, combined with a Fourier spectral discretization in space. The spatial domain is discretized on a $256 \times 256$ periodic grid, and the output time interval is $\Delta t = 0.05$.

The candidate library consists of all monomials in $u$ and $v$ up to total degree 3, together with the first- and second-order spatial derivatives of both variables, yielding a total of 20 candidate terms~\cite{deSilva2020PySINDy}:
\[
\left\{ u^i v^j \;\middle|\; i,j \ge 0,\ i+j \le 3 \right\}
\cup
\left\{ u_x, u_y, u_{xx}, u_{xy}, u_{yy}, v_x, v_y, v_{xx}, v_{xy}, v_{yy} \right\}.
\]
The required derivatives for constructing this library are estimated via automatic differentiation~\cite{Baydin2018ADSurvey,Xu2021DLPDE,Kaheman2022Automatic} on a neural-network surrogate (a fully connected network with five hidden layers of 50 neurons each using sinusoidal activation functions) trained on 7,500 points (0.114\% of the full dataset) obtained via Latin hypercube sampling~\cite{Baydin2018ADSurvey,Xu2021DLPDE,Kaheman2022Automatic}.

\begin{figure}[htbp!]
\centering
\includegraphics[width=1.00\textwidth]{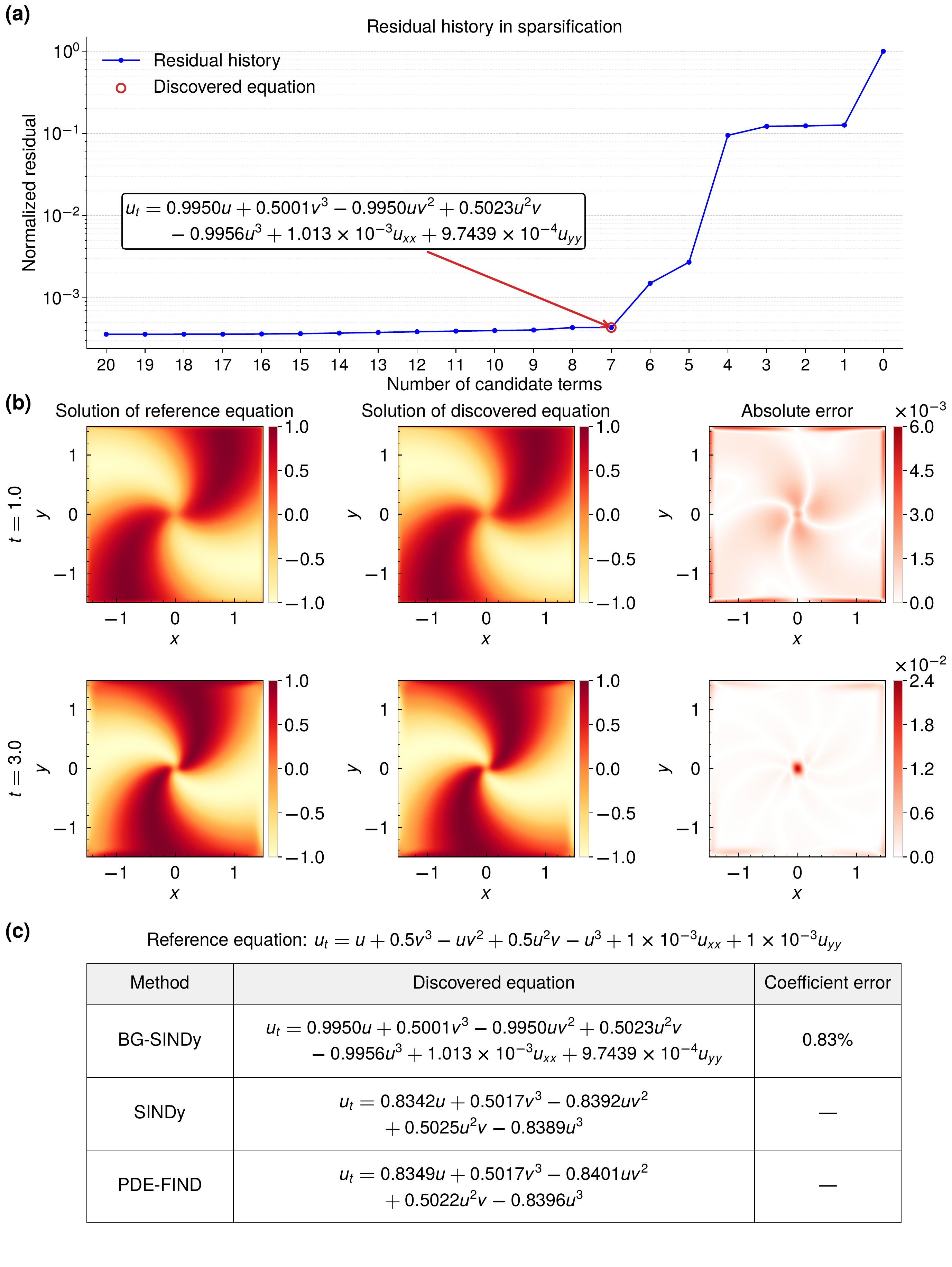}
\caption{Results for the 2D reaction--diffusion equation ($u$ component). 
(a) History of residual in the sparsification process. The red circle denotes the discovered equation.
(b) Comparison of solution snapshots from the reference and discovered equations at $t=1.0$ and $t=3.0$. The relative $L_2$ error, computed over the entire time domain, is $0.22\%$.
(c) Discovered equation by BG-SINDy and comparison with SINDy and PDE-FIND. BG-SINDy accurately recovers the reference equation, whereas SINDy and PDE-FIND fail to identify the equation structure.}
\label{fig:RD_u}
\end{figure}

\begin{figure}[htbp!]
\centering
\includegraphics[width=1.00\textwidth]{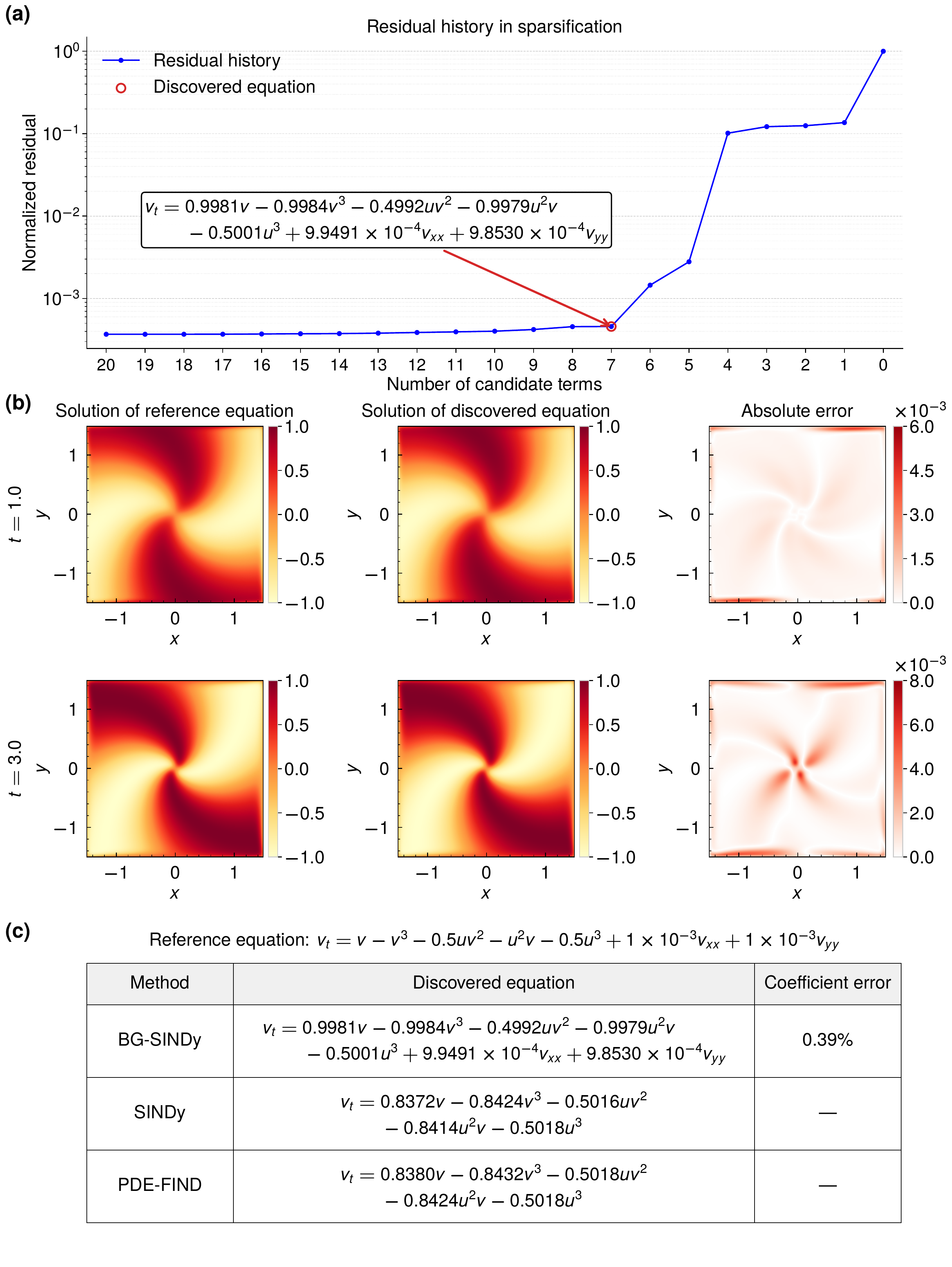}
\caption{Results for the 2D reaction--diffusion equation ($v$ component). 
(a) History of residual in the sparsification process. The red circle denotes the discovered equation.
(b) Comparison of solution snapshots from the reference and discovered equations at $t=1.0$ and $t=3.0$. The relative $L_2$ error, computed over the entire time domain, is $0.14\%$.
(c) Discovered equation by BG-SINDy and comparison with SINDy and PDE-FIND. BG-SINDy accurately recovers the reference equation, whereas SINDy and PDE-FIND fail to identify the equation structure.}
\label{fig:RD_v}
\end{figure}

The results for the $u$ and $v$ components are summarized in Fig.~\ref{fig:RD_u} and Fig~\ref{fig:RD_v}, respectively. For each component, Fig.~\ref{fig:RD_u}(a) and Fig.~\ref{fig:RD_v}(a) shows the evolution of the residual during the sparsification process, with the red circles indicating the selected models. Fig.~\ref{fig:RD_u}(b) and Fig.~\ref{fig:RD_v}(b) compares solution snapshots from the reference simulation and the discovered model at $t=1.0$ and $t=3.0$. The relative $L_2$ errors for the $u$ and $v$ components, evaluated over the entire time domain, are $0.22\%$ and $0.14\%$, respectively. As shown in Fig.~\ref{fig:RD_u}(c) and Fig.~\ref{fig:RD_v}(c), BG-SINDy accurately recovers the reference equations for both components, whereas standard SINDy and PDE-FIND fail to identify the correct equation structures. This example demonstrates the scalability of BG-SINDy to two-dimensional coupled systems.

\section{Conclusions~\label{sec:4}}
In this paper, we propose BG-SINDy, a data-driven framework for discovering nonlinear governing equations in multiscale systems with small-coefficient terms. Motivated by the principle of dominant balance, the method reformulates the $\ell_0$-constrained sparse regression (Eq.~\eqref{eq:SINDy}) as a term-level $\ell_{2,0}$-regularized formulation (Eq.~\eqref{eq:BG-SINDy}) and solves it using a progressive pruning strategy. Candidate terms are ranked according to their relative contributions to the equation balance and progressively pruned, with coefficients re-estimated via iterative least-squares regression, enabling the identification of dynamically essential terms even when their coefficients are small.

The performance of BG-SINDy is illustrated on a set of benchmark problems, including the KdV equation with a small dispersion coefficient, a modified Burgers equation with vanishing hyperviscosity, a modified KS equation with multiple small-coefficient terms, and a two-dimensional reaction--diffusion system. In these examples, the method is able to recover the underlying partial differential equations, including terms that are often difficult to identify using conventional sparse-regression approaches. This behavior is associated with a balance-aware term selection strategy, which differs from standard approaches that rely on coefficient magnitude and instead evaluates the contributions of candidate terms to the governing dynamics.

More broadly, incorporating dominant-balance concepts into equation-discovery algorithms may provide a useful perspective for constructing interpretable models of complex physical systems. Future work will explore extensions to other equation-discovery frameworks, including conventional sparse regression and symbolic regression, as well as integration with deep learning approaches for learning governing equations and surrogate models from sparse or noisy data. These directions are particularly relevant for data-driven modeling of challenging multiscale phenomena, such as non-Newtonian constitutive relations~\cite{ZHAO2018116} and subgrid-scale closures in turbulence~\cite{Laure2020Oceanturbulence,BenjaminSanderse2025turbulencereview,choi2026turbulence,Jakhar2026turbulence}.

\section*{Declaration of competing interest}

The authors declare that they have no known competing financial interests or personal relationships that could have
appeared to influence the work reported in this paper.

\section*{Acknowledgements}
This work was supported by the National Natural Science Foundation of China (NSFC) under the Excellence Research Group Program ``Multiscale Problems in Nonlinear Mechanics'' (Grant No. 12588201). 
Lei Zhang was also supported by the NSFC under Grant No. 12202451. Zhenhua Dang gratefully acknowledges Zihao Yang and Junfeng Zhao at Northwestern Polytechnical University for helpful discussions and support. Long Wang also acknowledges valuable discussions with Fengshun Zhang at the Institute of Mechanics, Chinese Academy of Sciences.

\newpage
\bibliographystyle{elsarticle-num}
\bibliography{cas-refs}

\end{document}